\documentclass[10pt,twocolumn,letterpaper]{article}

\usepackage{cvpr}
\usepackage{times}
\usepackage{epsfig}
\usepackage{graphicx}
\usepackage{amsmath}
\usepackage{amssymb}

\usepackage{algorithm}
\usepackage[noend]{algorithmic}
\usepackage{multirow}
\usepackage{hhline}
\usepackage{sidecap}

\usepackage[labelformat=simple]{subcaption}
\usepackage{tabularx}
\usepackage{enumitem}

\usepackage{xcolor}
\usepackage{booktabs}
\usepackage{capt-of}

\allowdisplaybreaks

\newcolumntype{C}{>{\centering}X}
\newcolumntype{L}{>{\raggedright\arraybackslash}X}

\usepackage[pagebackref=true,breaklinks=true,letterpaper=true,colorlinks,bookmarks=false]{hyperref}

\cvprfinalcopy 

\def\cvprPaperID{3020} 

\ifcvprfinal\fi

\makeatletter
\def\ps@myheadings{%
    \let\@oddfoot\@empty\let\@evenfoot\@empty
    \def\@evenhead{\thepage\hfil\slshape\leftmark}%
    \def\@oddhead{{\slshape\rightmark}\hfil\thepage}%
    \let\@mkboth\@gobbletwo
    \let\sectionmark\@gobble
    \let\subsectionmark\@gobble
    }
\renewcommand\maketitle{\par
  \begingroup
    \renewcommand\thefootnote{\@fnsymbol\c@footnote}%
    \def\@makefnmark{\rlap{\@textsuperscript{\normalfont\@thefnmark}}}%
    \long\def\@makefntext##1{\parindent 1em\noindent
            \hb@xt@1.8em{%
                \hss\@textsuperscript{\normalfont\@thefnmark}}##1}%
    \if@twocolumn
      \ifnum \col@number=\@ne
        \@maketitle
      \else
        \twocolumn[\@maketitle]%
      \fi
    \else
      \newpage
      \global\@topnum\z@   
      \@maketitle
    \fi
    \thispagestyle{plain}\@thanks
  \endgroup
  \setcounter{footnote}{0}%
}
\makeatother

\begin{document}

\title{Single Image Depth Estimation Trained via Depth from Defocus Cues}

\author{Shir Gur\\
Tel Aviv University\\
{\tt\small shir.gur@cs.tau.ac.il}
\and
Lior Wolf\\
Facebook AI Research and Tel Aviv University\\
{\tt\small wolf@cs.tau.ac.il}
}

\maketitle
\thispagestyle{empty}

\begin{abstract}
Estimating depth from a single RGB images is a fundamental task in computer vision, which is most directly solved using supervised deep learning. In the field of unsupervised learning of depth from a single RGB image, depth is not given explicitly. Existing work in the field receives either a stereo pair, a monocular video, or multiple views, and, using losses that are based on structure-from-motion, trains a depth estimation network. In this work, we rely, instead of different views, on depth from focus cues. Learning is based on a novel Point Spread Function convolutional layer, which applies location specific kernels that arise from the Circle-Of-Confusion in each image location. We evaluate our method on data derived from five common datasets for depth estimation and lightfield images, and present results that are on par with supervised methods on KITTI and Make3D datasets and outperform unsupervised learning approaches. Since the phenomenon of depth from defocus is not dataset specific, we hypothesize that learning based on it would overfit less to the specific content in each dataset. Our experiments show that this is indeed the case, and an estimator learned on one dataset using our method provides better results on other datasets, than the directly supervised methods.
\end{abstract}

\section{Introduction}
In classical computer vision, many depth cues were used in order to recover depth from a given set of images. These shape from X methods include structure-from-motion, which is based on multi-view geometry, shape from structured light, in which the known light source plays the role of an additional view, shape from shadow, and most relevant to our work, shape from defocus. In machine learning based computer vision, the interest has mostly shifted into depth from a single image, treating the problem as a multivariant image-to-depth regression problem, with an additional emphasis on using deep learning.

Learning depth from a single image consists of two forms. There are supervised methods, in which the target information (the depth) is explicitly given, and unsupervised methods, in which the depth information is given implicitly. The most common approach in unsupervised learning is to provide the learning algorithm with stereo pairs or other forms of multiple views~\cite{wang2018learning,yin2018geonet}. In these methods, the training set consists of multiple scenes, where for each scene, we are given a set of views. The output of the method, similar to the supervised case, is a function that given a single image, estimates depth at every point. 

In this work, we rely, instead of multiple view geometry, on shape from defocus. The input to our method, during training, is an all-in-focus image and one or more focused images of the same scene from the same viewing point. The algorithm then learns a regression function, which, given an all-in-focus image, estimates depth by reconstructing the given focused images. In classical computer vision, research in this area led to a variety of applications~\cite{zhuo2011defocus, tang2017depth, surh2017noise}, such as estimating depth from mobile phone images~\cite{suwajanakorn2015depth}. 
A deep learning based approach was presented by Anwar \etal~\cite{anwar2017depth} who employ synthetic focus images in supervised depth learning, and an aperture supervision depth learning by Srinivasan \etal~\cite{srinivasan2018aperture}, who employ lightfield images in the same way we use defocus images.

Our method relies on a novel Point Spread Function (PSF) layer, which preforms a local operation over an image, with a location dependent kernel which is computed ``on-the-fly'', according to the estimated parameters of the PSF at each location. More specifically, the layer receives three inputs: an all-in-focus image, estimated depth-map and camera parameters, and outputs an image at one specific focus. This image is then compared to the training images to compute a loss. Both the forward and backward operations of the layer are efficiently computed using a dedicated CUDA kernel. 
This layer is then used as part of a novel architecture, combining the successful ASPP architecture~\cite{chen2018encoder,fu2018deep}.
To improve the ASPP block, we add dense connections~\cite{huang2017densely}, followed by self-attention~\cite{zhang2018self}.

We evaluate our method on all relevant benchmarks we were able to obtain. These include the flower lightfield dataset and the multifocus indoor and outdoor scene dataset, for which we compare the ability to generate unseen focus images with other methods. We also evaluate on the KITTI, NYU, and Make3D, which are monocular depth estimation datasets. In all cases, we show an improved performance in comparison to methods with a similar level of supervision, and performance that is on par with the best directly supervised methods on KITTI and Make3D datasets. We note that our method uses focus cues for depth estimation, hence the task of defocusing for itself is not evaluated.

When learning depth from a single image, the most dominant cue is often the content of the image. For example, in street view images one can obtain a good estimate of the depth based on the type of object (sidewalk, road, building, car) and its location in the image. We hypothesize that when learning from focus data, the role of local image statistics becomes more dominant, and that these image statistics are more global between different visual domains. We therefore conduct experiments in which a depth estimator trained on one dataset is evaluated on another. Our experiments show a clear advantage to our method, in comparison to the state-of-the-art supervised monocular method of~\cite{fu2018deep}.

\section{Related Work}
\paragraph{Learning based monocular depth estimation}
In monocular depth estimation, a single image is given as input, and the output is the predicted depth associated with that image. Supervised training methods learn from the ground truth depth directly and the so-called unsupervised methods employ other data cues, such as stereo image pairs. 
One of the first methods in the field was presented by Saxena \etal~\cite{saxena2006learning}, applying supervised learning and proposed a patch-based model and Markov Random Field (MRF). Following this work, a variety of approaches had been presented using hand crafted representations~\cite{saxena2009make3d, ladicky2014pulling, ranftl2016dense, furukawa2017depth}. Recent methods use convolutional neural networks (CNN), starting from learning features for a conditional random field (CRF) model as in Liu \etal~\cite{liu2016learning}, to learning end-to-end CNN models refined by CRFs, as in~\cite{cao2017estimating, xu2017multi}. 

Many models employ an autoencoder structure~\cite{eigen2015predicting, garg2016unsupervised, kuznietsov2017semi,laina2016deeper, xie2016deep3d,fu2018deep}, with an added advantage to very deep networks that employ ResNets~\cite{he2016deep}. Eigen \etal~\cite{eigen2014depth,eigen2015predicting} showed that using multi-scaled depth predictions helps with the decrease in spatial resolution, which happened in the encoder model, and improves depth estimation. Other work uses different loss for regression, such as the reversed Huber~\cite{owen2007robust} used by Laina \etal~\cite{laina2016deeper} to lower the smoothness effect of the $L_2$ norm, and the recent work by Fu \etal~\cite{fu2018deep} who uses ordinal regression for each pixel with their spacing-increasing discretization (SID) strategy to discretize depth.

\noindent\textbf{Unsupervised depth estimation}\quad 
Modern methods for unsupervised depth estimation have relied on the geometry of the scene, Garg \etal~\cite{garg2016unsupervised} for example, proposed using stereo pairs for learning, introducing the differentiable inverse warping. Godard \etal~\cite{godard2017unsupervised} added the Left-Right consistency constraint to the loss function, exploiting another geometrical cue. Zhou \etal~\cite{zhou2017unsupervised} learned, in addition the ego-motion of the scene, and GeoNet~\cite{yin2018geonet} also used the optical flow of the scene. Wang \etal~\cite{wang2018learning} recently showed that using direct visual odometry along with depth normalization substantially improves performance on prediction.

\smallskip\noindent\textbf{Depth from focus/defocus}\quad 
The difference between depth from focus and depth from defocus is that, in the first case, camera parameters can be changed during the depth estimation process. In the second case, this is not allowed. Unlike the motion based methods above, these methods obtain depth using the structure of the optical geometry of the lens and light ray, as described in Sec.~\ref{sec:DfD}. Work in this field mainly focuses on analytical techniques. 
Zhuo \etal~\cite{zhuo2011defocus} for example, estimated the amount of spatially varying defocus blur at edge locations. The use of Coded Aperture had been proposed by~\cite{levin2007image, veeraraghavan2007dappled, sellent2014side} to improve depth estimation. 
Later work in this field, such as Suwajanakorn \etal~\cite{suwajanakorn2015depth}, Tang \etal~\cite{tang2017depth} and Surh \etal~\cite{surh2017noise} employed focal stacks --- sets of images of the same scene with different focus distances --- and estimated depth based on a variety of blurring models, such as the Ring Difference Filter~\cite{surh2017noise}. 
These methods first reconstruct an all-in-focus image and then optimize a depth map that best explains the re-rendering of the focal stack images out of the all-in-focus image.

There are not many deep learning works in the field. Srinivasan \etal~\cite{srinivasan2018aperture} presented a new lightfield dataset of flower images. They used the ground truth lightfield images to render focused images and employed a regression model to estimate depth from defocus by reconstruction of the rendered focused images.{\color{black}While Srinivasan \etal~\cite{srinivasan2018aperture} did not compare to other RGB-D datasets~\cite{Geiger2013IJRR,saxena2006learning, saxena2007learning,Silberman:ECCV12}, their method can take as input any all-in-focus image. We evaluate~\cite{srinivasan2018aperture} rendering process using our network on the KITTI dataset.}
Anwar \etal~\cite{anwar2017depth} utilized the provided depth of those datasets to integrate focus rendering within a fully supervised depth learning scheme.

\section{Differentiable Optical Model}
We review the relevant optical geometry on which our PSF layer relies and then move to the layer itself.

\subsection{Depth From Defocus}
\label{sec:DfD}
\begin{figure}
    \centering
    \begin{subfigure}[b]{0.15\textwidth}
    	\centering
		\includegraphics[width=\linewidth]{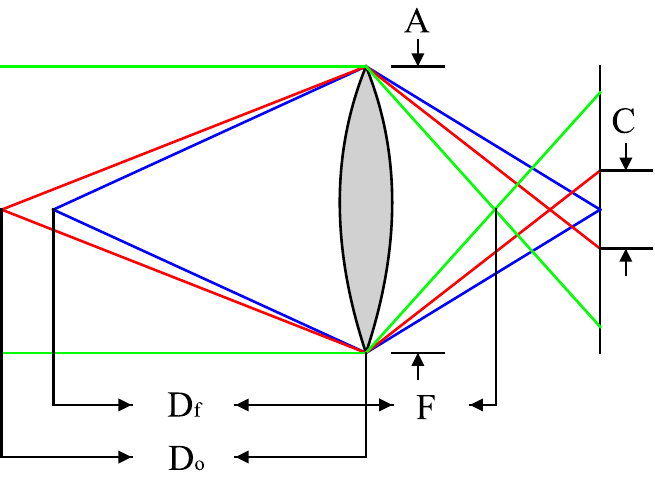}
        \caption{Lens illustration}
        \label{fig:lens}
    \end{subfigure}
    \hfill
    \begin{subfigure}[b]{0.15\textwidth}
    	\centering
		\includegraphics[width=\linewidth]{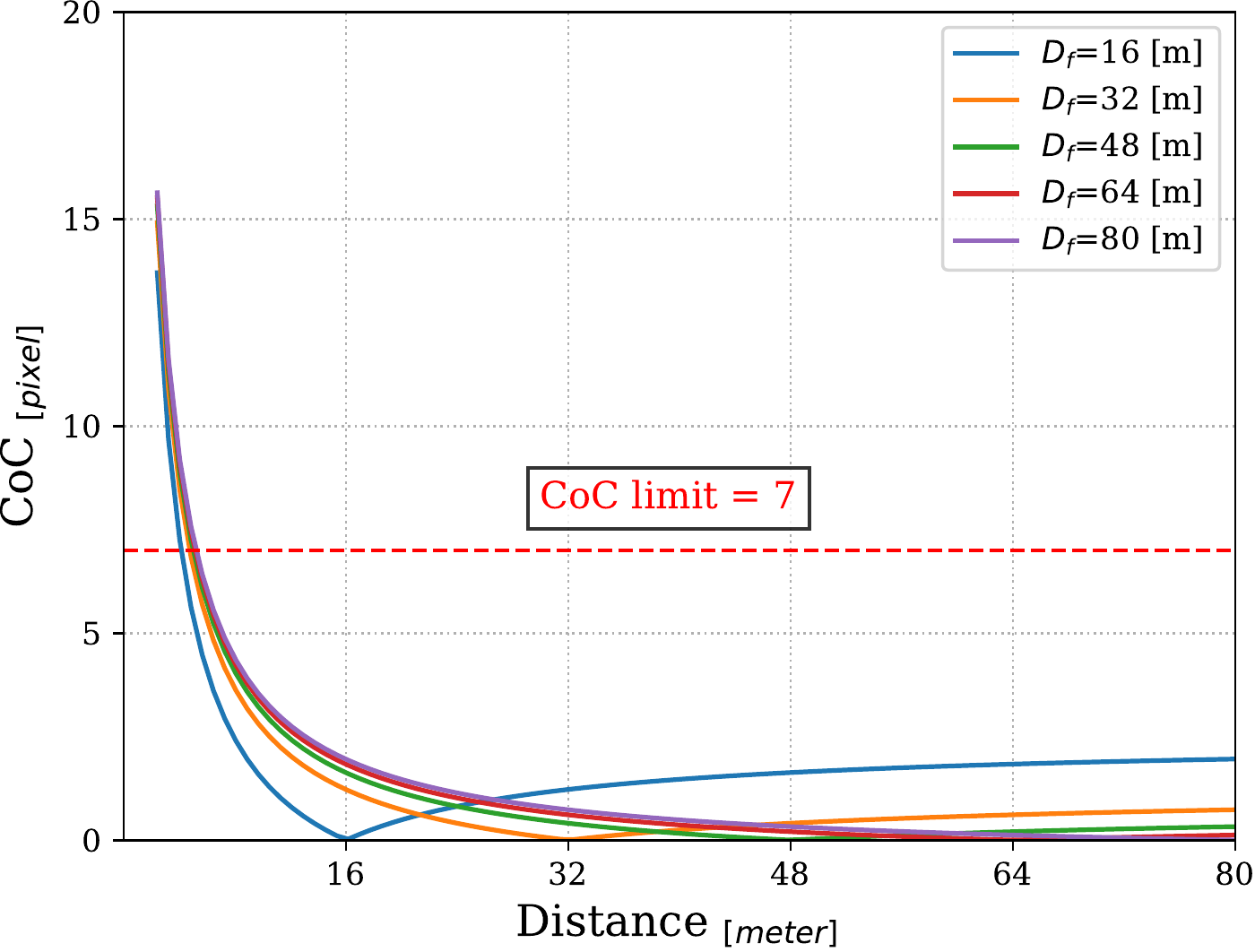}
        \caption{CoC - KITTI}
        \label{fig:CoC_KITTI}
    \end{subfigure}
    \hfill
    \begin{subfigure}[b]{0.15\textwidth}
    	\centering
		\includegraphics[trim=0 .31cm 2.96cm 0cm, clip,width=0.75\linewidth]{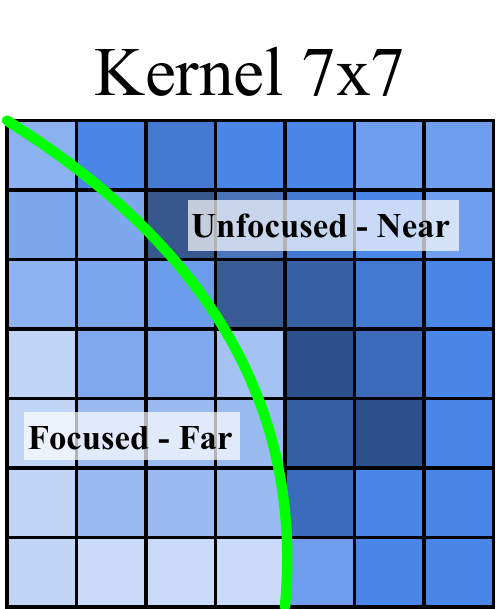}
        \caption{CoC - KITTI}
        \label{fig:psf}
    \end{subfigure}
    \caption{(a) Illustration of lens principles. Blue beams represent an object in focus. Red beams represent an object further away and out of focus. See text for symbol definitions. (b) CoC diameter \wrt object distance as seen in KITTI. Camera settings are: $N=2.8$, $F=35$, and $s=2$. {\color{black}(c) Sample blur kernel. Green line represents depth edge, Blue colors represent the relative blur contribution \wrt CoC.}}
\end{figure}
Depth from focus methods are mostly based on the thin-lens model and geometry, as shown in Fig.~\ref{fig:lens}. The figure illustrates light rays trajectories and the blurring effect made by out-of-focus objects. The plane of focus is defined 
such that light rays emerging from it towards the lens fall at the same point on the camera sensor plane. An object is said to be in focus, if its distance from the lens falls inside the camera's depth-of-field (DoF), which is the distance about the plane of focus where objects appear acceptably sharp by the human eye. Objects outside the DoF appear blurred on the image plane, an effect caused by the spread of light rays coming from the unfocused objects and forming what is called the ``Circle-Of-Confusion'' (CoC), as marked by C in Fig.~\ref{fig:lens}. 
In this paper, we will use the following terminology: an \textit{all-in-focus} image is an image where all objects appear in focus, and a \textit{focused} image is one where blurring effects caused by the lens configuration are observed.

In this model, we consider the following parameters to describe a specific camera: focal-length $F$, which is the distance between the lens plane and the point where initially parallel rays are brought to a focus, aperture $A$, which is the diameter of the lens (or an opening through which light travels), and the plane of focus $D_f$ (or focus distance), which is the distance between the lens plane and the plane where all points are in focus. 
Following the thin-lens model, we define the size of blur, \ie., the diameter of the CoC, which we denote as $C_{mm}$, according to the following equation:
\begin{align}
	\label{eq:coc}
	C_{mm} = A \frac{|D_o - D_f|}{D_o} \frac{F}{D_f - F}
\end{align}
where $D_o$ is the distance between an object to the lens plane, and $A=F/N$ where $N$ is what is known as the f-number of the camera.
While CoC is usually measured in millimeters ($C_{mm}$), we transform its size to pixels by considering a camera pixel-size of $p=5.6\mu m$ as in~\cite{Carvalho2018eccv3drw}, and a camera output scale $s$, which is the ratio between sensor size and output image size. The final CoC size in pixels $C$ is computed as follows:
\begin{align}
	C = \frac{C_{mm}}{p \cdot s}.
\end{align}

The CoC is directly related to the depth, as illustrated in Fig.~\ref{fig:CoC_KITTI}, where each line represents a different focus distance $D_f$. As can be seen, the relation is not one-to-one and will cause ambiguity in depth estimation. 
Moreover, different camera settings are required for different scenes in terms of the scene's maximum depth, \ie for KITTI, we consider maximum depth of 80 meters, and 10 meters for NYU. We also consider a constant f-number of $N=2.8$ and a different focal-length for all datasets, in order to lower depth ambiguity by lowering the DoF range (see Sec.~\ref{sec:depth_ambiguity} for more details).

We now refer to one more measurement named CoC-limit, defined as the largest blur spot that will still be perceived by the human eye as a point, when viewed on a final image from a standard viewing distance. The CoC-limit also limits the kernel size used for rendering and is, therefore, highly influential on the run time (bigger kernels lead to more computations). We employ a kernel of size $7 \times 7$, which reflects a standard CoC-limit of $0.061mm$.

In this work, following~\cite{suwajanakorn2015depth, tang2017depth}, we consider the blur model to be a disc-shaped point spread function (PSF), modeled by a Gaussian kernel with radius $r = C/2$ and kernel's location indices $u, v$:
\begin{align}
	\label{eq:gaus}
	G(u, v, r) = \frac{1}{2\pi r^2}\exp\bigg(-\bigg(\frac{u^2+ v^2}{2r^2}\bigg)\bigg)
\end{align}
Because we work in pixel space, if the diameter is less then one pixel ($C<1$), we ignore the blurring effect.

According to the above formulation, a focused image can be generated from an all-in-focus image and depth-map, as commonly done in graphics rendering. Let $I$ be an all-in-focus image and $J$ be a rendered focused image derived from depth-map $D_o$, CoC-map $C$, camera parameters $A$, $F$ and $D_f$, we define $J$ as follows:
\begin{align}
	\label{eq:main_f}
	\mathcal{F}_{x,y}(u, v) &= \frac{2}{\pi C_{x,y}^2}\exp\bigg(-2\bigg(\frac{u^2+ v^2}{C_{x,y}^2}\bigg)\bigg) \\
    \label{eq:main}
	J_{x, y} :&= (I \circledast F) \\
    \nonumber
    &= \frac{\int\limits_{u,v \in \Omega} I_{x-u, y-v}\mathcal{F}_{x-u, y-v}(u, v)du dv}{\int\limits_{u',v' \in \Omega} \mathcal{F}_{x-u', y-v'}(u', v')du' dv'},
\end{align}
where $\Omega$ is an offsets set related to a kernel of size $m \times m$:
\begin{align}
	\Omega := \left\lbrace (u,v):\;u,v\in\bigg[-\frac{m}{2},\dots,0,\dots,\frac{m}{2}\bigg]\in\mathbb{N} \right\rbrace 
\end{align}
We denote by $\circledast$ the convolution operation with a functional kernel $\mathcal{F}$, by $(x,y)$ the image location indices, and by $(u,v)$ the offset indices bounded by the kernel size.

Based on Eq.~\ref{eq:main}, given a set of focused images of the same scene, one may optimize a model to predict the all-in-focus image and the depth map. Alternatively, given a focused image and its correspondent all-in-focus image, we predict the scene depth by reconstructing the focused image.

{\color{black}While~\cite{srinivasan2018aperture} uses a weighted sum of disk kernels to render blur, our blur kernel is a Gaussian composition of different blur contributions from all neighbors (Eq.~\ref{eq:main}) where each kernel coefficient is calculated by a Gaussian function \wrt a different estimated CoC, as illustrated in Fig.~\ref{fig:psf}.}

\subsection{The PSF Convolutional layer}
\label{sec:psf}
The PSF layer we employ can be seen as a particular case of the locally connected layers of~\cite{taigman2014deepface}, with a few differences: first, in the PSF layer, the same operator is applied across all channels, while in the locally-connected layer, as well as in conventional layers (excluding depth-convolution~\cite{chollet2017xception}), the local operator varies between the input channels.
Additionally, The PSF layer does not sum the outcomes, and returns the same number of channels in the output tensor as in the input tensor.

The PSF convolutional layer, designed for the task of Depth from Defocus (DfD), is based on Eq.~\ref{eq:main}, where kernels vary between locations and are calculated ``on-the-fly'', according to function $\mathcal{F}$, which is defined in Eq.~\ref{eq:main_f}. The kernel is, therefore, a local function of the object's distance, with a blur kernel applied to out-of-focus pixels.
The layer takes as input an all-in-focus image $I$, depth-map $D_o$ and the camera parameters vector $\rho$, which contains the aperture $A$, the focal length $F$ and the focal depth $D_f$. The layer then outputs a focused image $J$. As mentioned before, we fix the near and far distance limits to fit each dataset and use the fixed pixel size mentioned above. The rendering process begins by first calculating the CoC-map $C$ according to Eq.~\ref{eq:coc}, and then applying the functional kernel convolution defined in Eq.~\ref{eq:main}. We implement the following operation in CUDA and compute its derivative as follows:

\begin{align}
  \bigg(\frac{\partial J_{s,t}}{\partial I_{x,y}}\bigg) &= \frac{\mathcal{F}_{x,y}(u,v)}{\int\limits_{u',v' \in \Omega} \mathcal{F}_{s-u',t-v'}(u', v')du' dv'}\label{eq:6}\\
  \bigg(\frac{\partial J_{s,t}}{\partial C_{x,y}}\bigg) &= \frac{\xi_{x,y}(u, v) (I_{x,y}-J_{s,t})\mathcal{F}_{x,y}(u, v)}{\int_{u',v' \in \Omega} \mathcal{F}_{s-u',t-v'}(u', v')du' dv'}\label{eq:7}\\
  \xi_{x,y}(u, v) :&= \frac{4(u^2+ v^2) - 2C_{x,y}^2}{C_{x,y}^3}
\end{align}

A detailed explanation of the forward and backward pass is provided in the supplementary material.

\section{Approach}
\label{approach}
In this section, we describe the training method and the model architecture, which extends the ASPP architecture to include both self-attention and dense connections. We then describe the training procedure.

\subsection{General Architecture and the Training Loss}
Let $J$ be a (real-world) focused version of $I$, and $\bar{J}$ be a predicted focused version of $I$. We train a regression model to minimize the reconstruction loss of $J$ and $\bar{J}$.

We define two networks, $f$ and $g$, for depth estimation and focus rendering respectively. While $f$ is learned, $g$ implements Eq.~\ref{eq:main_f} and~\ref{eq:main}. Both networks take part in the loss, and backpropagation through $g$ is performed using Eq.~\ref{eq:6},~\ref{eq:7}.

The learned network $f$ is applied to an all-in-focus image $I$ and returns a predicted depth $\bar{D}_o= f(I)$.
The fixed network $g$ consists of the PSF layer, as described in Sec.~\ref{sec:psf}. It takes as input an all-in-focus $I$, a depth (estimated or not) $D_o$ and the camera parameters vector $\rho$. It outputs $J=g(I,D_o,\rho)$, which is a focused version of $I$ according to depth $D_o$ and camera parameters $\rho$.
We distinguish between a rendered focus image from ground truth depth $D_o$ which we denote as $J$ (also used for real focused imaged), and rendered focused image from predicted depth $\bar{D}_o$, which we denote as $\bar{J}=g(I,\bar{D}_o,\rho)$.

\begin{figure}[t]
	\centering
	\includegraphics[width=\linewidth]{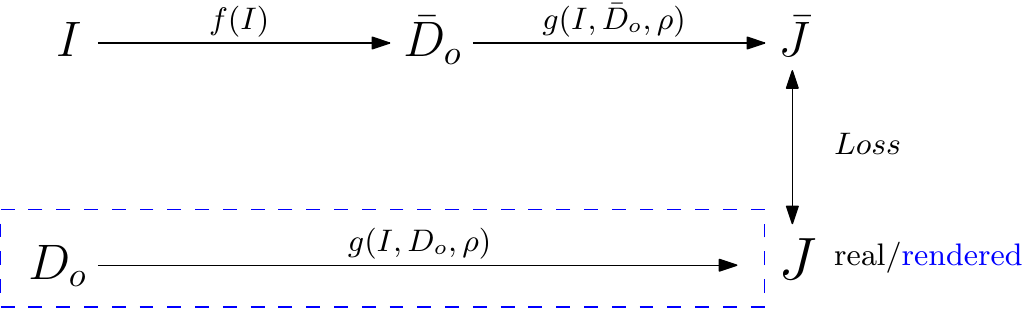}
	\caption{Training scheme. Blue region represents the rendering branch, which is used for depth-based datasets.}
	\label{fig:model_flow}
\end{figure}

The training procedure has two cases, training with real data or on generated data, depending on the training dataset at hand. In both cases, training is performed end-to-end by running $f$ and $g$ sequentially. 
First, $f$ is applied to an all-in-focus image $I$ and outputs the predicted depth-map $\bar{D}_o$. Using this map, the all-in-focus image and camera parameters $\rho$, $g$ renders the predicted focused image $\bar{J}$.
A reconstruction error is then applied with $J$ and $\bar{J}$, where for the case of depth-based datasets, we render the training focused images $J$, according to ground truth depth-map $D_o$ and camera specifications $\rho$.
Fig.~\ref{fig:model_flow} shows the training scheme, where the blue dashed rectangle illustrates the second case, where $J$ is rendered from the ground truth depth.

In the first case, since we compare with the work of~\cite{srinivasan2018aperture}, we use a single focused image during training, although more can be used. In the second case, we compare with fully supervised methods, that benefit from a direct access to the depth information, and we report results for 1, 2, 6 and 10 rendered focused images.

\noindent\textbf{Training loss}\quad
We first consider the reconstruction loss and the depth smoothness~\cite{wang2004image, godard2017unsupervised} \wrt the input image $I$, the predicted focused image $\bar{J}$, the focused image $J$, and the estimated depth map $\bar{D}_o$:
\begin{gather}
    \mathcal{L}_{rec} = \frac{1}{N}\sum{}{}\alpha\frac{1 - SSIM(\bar{J}, J)}{2} + (1 - \alpha)\|\bar{J}-J\|_1\\
    \mathcal{L}_{smooth} = \frac{1}{N}\sum{}{}|\partial_x \bar{D}_o|e^{-|\partial_x I|} + |\partial_y \bar{D}_o|e^{-|\partial_y I|}
\end{gather}
where $SSIM$ is the Structural Similarity measure~\cite{wang2004image}, and $\alpha$ controls the balance \wrt to $L_1$ loss.

The reconstruction loss above does not take into account the blurriness in some parts of image $J$, which arise from regions that are out of focus. We, therefore, add a sharpness measure $S(I)$ similar to~\cite{madhavi2011all}, which considers the sharpness of each pixel. It contains three parts: (i) the image \textit{Laplacian} 
$\Delta I := \partial_x^2I + \partial_y^2I$, 
(ii) the image \textit{Contrast Visibility}
$C(I) := \bigg|\frac{I - \mu_I}{\mu_I}\bigg|$
, and (iii) the image \textit{Variance}
$V(I) := (I - \mu_I)^2$,
where \(\mu_I\) is the average pixel value in a window of size $7 \times 7$ pixels. The sharpness measure is given by $S(I) = -\Delta I - C(I) - V(I)$, and the loss term is:
\begin{align}
    \mathcal{L}_{sharp} &= \|S(\hat{J}) - S(J)\|_1.
\end{align}
The final loss term is then:
\begin{align}
	Loss = \lambda_1\mathcal{L}_{rec} + \lambda_2\mathcal{L}_{smooth} + \lambda_3\mathcal{L}_{sharp}
\end{align}
For all experiments, we set  $\lambda_1=1, \lambda_2=10^{-3}, \lambda_3=10^{-1}$.

\subsection{Model Architecture}
\begin{figure}[t]
	\centering
	\includegraphics[width=\linewidth]{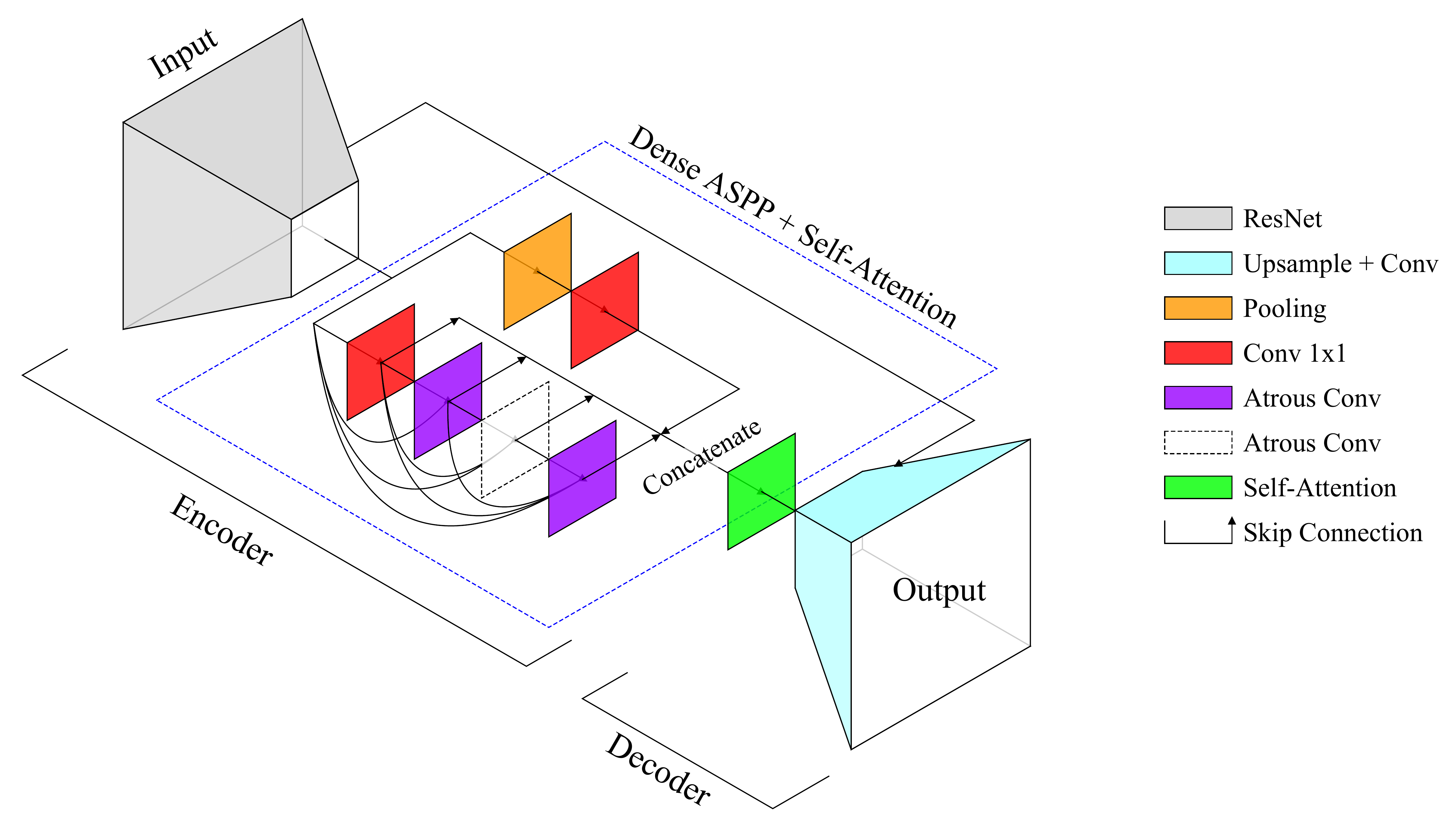}
	\caption{Dense ASPP with an added attention block.}
	\label{fig:arch}
\end{figure}

Our network $f$ is illustrated in Fig.~\ref{fig:arch}. It consists of an encoder-decoder architecture, where we rely on the DeepLabV3+~\cite{chen2017rethinking, chen2018encoder} model, which was found to be effective for semantic segmentation and depth estimation tasks~\cite{fu2018deep}. The encoder has two parts: a ResNet~\cite{he2016deep} backbone and a subsequent Atrous Spatial Pyramid Pooling (ASPP) module. Unlike~\cite{fu2018deep}, we do not employ a pretrained ResNet and learn it end-to-end.

The Atrous convolutions (also called dilated convolutions) add padding between kernel cells to enlarge the receptive field from earlier layers, while keeping the weight size constant. ASPP contains several parallel Atrous convolutions with different dilations. As advised in~\cite{chen2018encoder}, we also replace all pooling layers of the encoder with convolution layers with an appropriate stride. 

The loss is computed in the highest resolution, to support higher quality outputs. However, to comply with GPU memory constraints, the network takes as an input, a downsampled image of half the original size. The network's output is then upsampled to the original image size.

\noindent\textbf{Dense ASPP with Self-Attention}\quad 
The original ASPP consists of three or more independent layers - average pooling followed by $1 \times 1$ convolution, $1 \times 1$ convolution, and four Atrous layers. Each convolution layer has 256 channels and the four outputs of these layers, along with the pool+conv layer are concatenated together to form a tensor with channel size $C=1280$.
We propose two additional modifications from different parts of the literature: dense connections~\cite{huang2017densely} and self attention~\cite{zhang2018self}. 

We add dense connections between the $1 \times 1$ convolution and all Atrous convolution layers of the ASPP module, sequentially connecting all layers from smallest to the largest dilation layer. Each layer, therefore, receives as the input tensor not just the output of the previous layer, but the concatenation of the output tensors of all preceding layers. This is illustrated as the skip connection arrows in Fig.~\ref{fig:arch}.

Self-Attention aims to integrate local features with their global dependencies, and as shown in previous work~\cite{zhang2018self, fu2018dual}, it improve results in image segmentation and generation. Our implementation is based on~\cite{fu2018dual} dual-attention.

The decoder part of $f$ consists of three upsampling blocks, each having three convolution layers followed by bilinear upsampling. A skip connection from a low level layer of the backbone is concatenated with the input of the second block. The output of decoder is the predicted depth.

\section{Experiments}
We divide our experiments into two types, DoF supervision and DoF supervision from rendered data, as mentioned in the previous section. 
We further experiment with cross domain evaluation, where we evaluate our method in comparison to the state-of-the-art supervised method~\cite{fu2018deep}. Here the models are trained on domain A and tested on domain B, denoted as $A \rightarrow B$.
We show that learning depth from focus cues, though not achieving better results than the supervised methods - but comparable with top methods in KITTI and Make3D datasets, achieves better generalization expressed by higher results in cross domain evaluation.

The network is trained on a single Titan-X Pascal GPUs with batch size of 3, using Adam for optimization with a learning rate of $2 \cdot 10^{-5}$ and weight decay of $4 \cdot 10^{-5}$. The dedicated CUDA implementation of the PSF layer runs x80 faster than the optimized pytorch implementation. 

The following five benchmarks are used:

\noindent\textbf{Lightfield dataset~\cite{srinivasan2018aperture}}\quad
The dataset contains lightfield flowers and plants images, taken with a Lytro Illum camera. From the lightfield images, we follow the procedure of~\cite{srinivasan2018aperture} to generate the all-in-focus and shallow DoF images, and split the dataset into 3143 and 300 images for train and test. 

\noindent\textbf{DSLR dataset~\cite{Carvalho2018eccv3drw}}\quad
This dataset contains 110 images and ground truth depth from indoor scenes, with 81 images for training and 29 images for testing, and 34 images from outdoor scenes without ground truth depth. Each scene is acquired with two camera apertures: $N=2.8$ and $N=8$, providing focused and all-in-focus images.

\noindent\textbf{KITTI~\cite{Geiger2013IJRR}}\quad
This benchmark contains RGB-D images taken in an outdoor environment at resolution of roughly \(370 \times 1226 \) which we refer to as the full resolution output size. The  train/test splits we employ follow Eigen \etal~\cite{eigen2014depth}, with 23,000 training images and 697 test images. The input depth-maps and images are cropped, according to~\cite{eigen2014depth} to obtain valid depth values, and resized to half-size.

\noindent\textbf{NYU DepthV2~\cite{Silberman:ECCV12}}\quad
This benchmark contains about 120K indoor RGB and depth images captured with a Microsoft Kinect. The datasets consists of 249 scenes for training and 215 scenes for testing. We report results on 654 test images from a small subset of 1449 aligned RGB-depth pairs, as done in previous work.

\noindent\textbf{Make3D~\cite{saxena2006learning, saxena2007learning}}\quad
The Make3D benchmark contains 534 RGB-depth pairs, split into 400 pairs for training and 134 for testing. The input images are provided at a high resolution, while the depth-maps are at low resolution. Therefore, data is resized to $460 \times 345$, as proposed by~\cite{saxena2006learning, saxena2007learning}. Following~\cite{saxena2006learning}, results are evaluated in two settings: $C1$ for depth cap of 0-70, and $C2$ for depth cap 0-80.

\subsection{Results}
\noindent\textbf{DoF supervision}\quad
We first report results on the Lightfield dataset dataset, which provides focused and all-in-focus image pairs with no ground truth depth. The performance is evaluated using the PSNR and SSIM measures. Our results are shown in Tab.~\ref{table:Flowers}. As can be seen, we significantly outperform the literature baselines provided by ~\cite{srinivasan2018aperture}. 

\begin{table}[t]
	\small
	\centering
	\begin{tabularx}{\linewidth}{Lccc}
    	\toprule
		Algorithm & Supervision & PSNR & SSIM \tabularnewline 
		\midrule
		Image Regression~\cite{srinivasan2018aperture} & DoF & 24.60 & 0.895 \tabularnewline
		Multi-View~\cite{srinivasan2018aperture} & DoF & 34.49 & 0.960 \tabularnewline
		Lightfield~\cite{srinivasan2018aperture} & DoF & 36.68 & 0.967 \tabularnewline
		Compositional~\cite{srinivasan2018aperture} & DoF & 36.90 & 0.966 \tabularnewline
		\textbf{Ours} & DoF & \textbf{38.33} & \textbf{0.979} \tabularnewline 
		\bottomrule
	\end{tabularx}
	\caption{Quantitative results on the Lightfield test set, reported as a mean value of PSNR and SSIM of the reconstructed focused image.}
	\label{table:Flowers}
\end{table}    

\noindent\textbf{Rendered DoF supervision}\quad
\label{renderedDoFSup}
\setlength{\abovecaptionskip}{4pt plus 3pt minus 2pt}
\setlength{\belowcaptionskip}{8pt plus 3pt minus 2pt}
\begin{figure*}[!ht]
	\centering
    \small
    \begingroup
    \setlength{\tabcolsep}{1pt} 
    \renewcommand{\arraystretch}{1} 
    \begin{tabular}{cccccc}
        Reference Image & Ground Truth & Wang~\cite{wang2018learning} & F2 & F6 & F10 \\

        \includegraphics[width=0.16\linewidth]{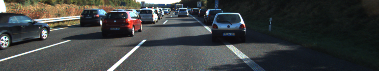} &
        \includegraphics[width=0.16\linewidth]{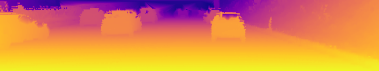} &
        \includegraphics[width=0.16\linewidth]{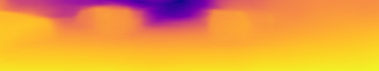} &
        \includegraphics[width=0.16\linewidth]{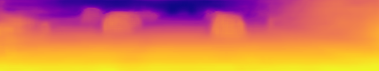} &
        \includegraphics[width=0.16\linewidth]{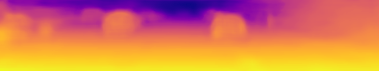} &
        \includegraphics[width=0.16\linewidth]{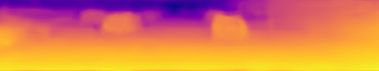}\\

        \includegraphics[width=0.16\linewidth]{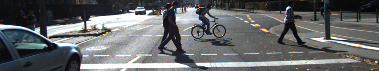} &
        \includegraphics[width=0.16\linewidth]{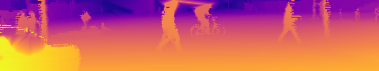} &
        \includegraphics[width=0.16\linewidth]{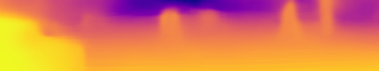} &
        \includegraphics[width=0.16\linewidth]{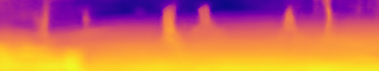} &
        \includegraphics[width=0.16\linewidth]{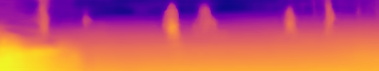} &
        \includegraphics[width=0.16\linewidth]{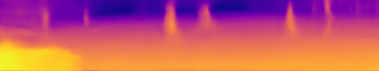}\\

        \includegraphics[width=0.16\linewidth]{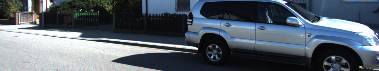} &
        \includegraphics[width=0.16\linewidth]{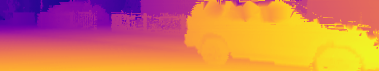} &
        \includegraphics[width=0.16\linewidth]{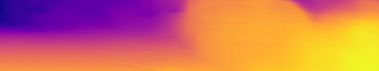} &
        \includegraphics[width=0.16\linewidth]{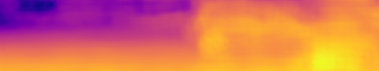} &
        \includegraphics[width=0.16\linewidth]{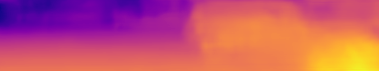} &
        \includegraphics[width=0.16\linewidth]{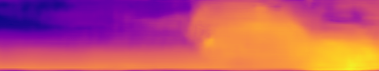}\\

        \includegraphics[width=0.16\linewidth]{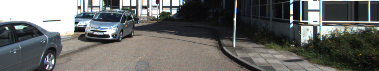} &
        \includegraphics[width=0.16\linewidth]{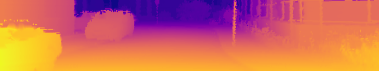} &
        \includegraphics[width=0.16\linewidth]{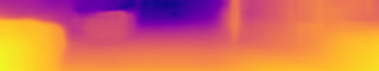} &
        \includegraphics[width=0.16\linewidth]{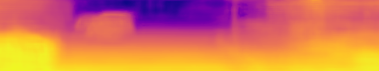} &
        \includegraphics[width=0.16\linewidth]{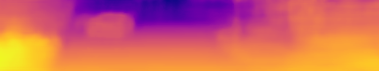} &
        \includegraphics[width=0.16\linewidth]{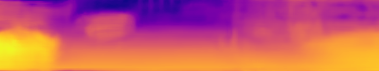}
    \end{tabular}
    \endgroup
    \captionof{figure}{\textbf{KITTI:} Qualitative results on the KITTI Eigen Split. All images are cropped to the valid depth region as proposed in~\cite{eigen2014depth}. From left to right, reference image and ground truth, Wang \etal~\cite{wang2018learning} and ours.}
	\label{figure:KITTI_Images}
	\begin{tabularx}{\linewidth}{Lcccccccc}
        \toprule
        Algorithm & Supervision & Abs Rel & Sq Rel & RMSE & RMSE log & \(\delta < 1.25\) & \(\delta < 1.25^2\) & \(\delta < 1.25^3\)\tabularnewline 
        \midrule
          Godard \etal ~\cite{godard2017unsupervised} & S & 0.148 & 1.344 & 5.927 & 0.247 & 0.803 & 0.922 & 0.964 \tabularnewline 
        Geonet-ResNet~\cite{yin2018geonet} & M & 0.155 & 1.296 & 5.857 & 0.233 & 0.793 & 0.931 & 0.973 \tabularnewline 
        Wang \etal~\cite{wang2018learning} & M & 0.151 & 1.257 & 5.583 & 0.228 & 0.810 & 0.936 & 0.974 \tabularnewline
        Godard \etal~\cite{godard2017unsupervised} & S(K+CS) & 0.114 & 0.898 & 4.935 & 0.206 & 0.861 & 0.949 & 0.976 \tabularnewline 
        \midrule
        Ours F1 & DoF & 0.141 & 1.473 & 5.187 & 0.221 & 0.846 & 0.953 & 0.981 \tabularnewline
        Ours F2 & DoF & 0.129 & 0.722 & 4.233 & 0.183 & 0.856 & 0.960 & 0.985\tabularnewline 
        Ours F6 & DoF & 0.114 & 0.671 & 4.144 & 0.172 & 0.867 & 0.963 & 0.987 \tabularnewline 
        Ours F10 & DoF & 0.110 & 0.666 & 4.186 & 0.168 & 0.880 & 0.966 & 0.988 \tabularnewline 
        \hline
        Liu \etal~\cite{liu2016learning} & Depth & 0.202 & 1.614 & 6.523 & 0.275 & 0.678 & 0.895 & 0.965 \tabularnewline 
        Kuznietsov \etal~\cite{kuznietsov2017semi} & Depth & 0.113 & 0.741 & 4.621 & 0.189 & 0.862 & 0.960 & 0.986 \tabularnewline 
        DORN \etal~\cite{fu2018deep} & Depth & 0.072 & 0.307 & 2.727 & 0.120 & 0.932 & 0.984 & 0.994 \tabularnewline 
        \bottomrule
	\end{tabularx}
	\captionof{table}{\textbf{KITTI:} Quantitative results on the KITTI Eigen split. \textbf{Top -} Unsupervised methods where `S' and `M' stands for stereo and video (monocular) supervision, and `K+CS' stands for training with the added data from the CityScapes dataset. \textbf{Middle -} Our method. \textbf{Bottom -} Supervised methods.}
	\label{table:KITTI}
    \begin{tabularx}{\linewidth}{Lccccccc}
        \toprule
        \multirow{2}{*}{Algorithm} & \multirow{2}{*}{Supervision} & \multicolumn{3}{c}{\(C1\)} & \multicolumn{3}{c}{\(C2\)} \tabularnewline
        & & Abs Rel & RMSE log\(_{10}\) & RMSE & Abs Rel & RMSE log\(_{10}\) & RMSE\tabularnewline
        \midrule
        Godard \etal~\cite{godard2017unsupervised} & S & 0.443 & 0.156 & 11.513 & - & - & - \tabularnewline
        Zhou \etal~\cite{zhou2017unsupervised} & MS & 0.383 & 0.478 & 10.470 & - & - & - \tabularnewline
        Wang \etal~\cite{wang2018learning} & MS & 0.387 & 0.204 & 8.090 & - & - & - \tabularnewline
        \midrule
        Ours F1 & DoF & 0.568 & 0.192 & 8.822 & 0.575 & 0.195 & 10.147 \tabularnewline 
        Ours F2 & DoF & 0.287 & 0.116 & 7.710 & 0.294 & 0.121 & 9.387 \tabularnewline 
        Ours F6 & DoF & 0.262 & 0.109 & 7.474 & 0.269 & 0.115 & 9.248 \tabularnewline 
        Ours F10 & DoF & 0.246 & 0.110 & 7.671 & 0.254 & 0.116 & 9.494 \tabularnewline 
        \midrule
        Li \etal~\cite{li2015depth} & Depth & 0.278 & 0.092 & 7.120 & 0.279 & 0.102 & 10.27 \tabularnewline 
        MS-CRF~\cite{xu2017multi} & Depth & 0.184 & 0.065 & 4.380 & 0.198 & - & 8.56 \tabularnewline 
        DORN~\cite{fu2018deep} & Depth & 0.157 & 0.062 & 3.970 & 0.162 & 0.067 & 7.32 \tabularnewline 
        \bottomrule
    \end{tabularx}
	\captionof{table}{\textbf{Make3D:} Quantitative results on Make3D~\cite{saxena2006learning, saxena2007learning} dataset. \textbf{Top -} Unsupervised methods where `S' and `M' stands for stereo and video (monocular) supervision. \textbf{Middle -} Our method. \textbf{Bottom -} Supervised methods.}
	\label{table:Make3D}
    \begin{tabularx}{\linewidth}{Lccccccc}
        \toprule
        Algorithm & Supervision & Abs Rel & RMSE log\(_{10}\) & RMSE & \(\delta < 1.25\) & \(\delta < 1.25^2\) & \(\delta < 1.25^3\)\tabularnewline 
        \midrule
        Ours F1 & DoF & 0.254 & 0.092 & 0.766 & 0.691 & 0.880 & 0.944 \tabularnewline 
        Ours F2 & DoF & 0.162 & 0.068 & 0.574 & 0.774 & 0.941 & 0.984 \tabularnewline 
        Ours F6 & DoF & 0.149 & 0.063 & 0.546 & 0.797 & 0.951 & 0.987 \tabularnewline 
        Ours F10 & DoF & 0.162 & 0.068 & 0.575 & 0.772 & 0.942 & 0.984 \tabularnewline 
        \midrule
        Li \etal~\cite{li2015depth} & Depth & 0.143 & 0.063 & 0.635 & 0.788 & 0.958 & 0.991 \tabularnewline 
        MS-CRF~\cite{xu2017multi} & Depth & 0.121 & 0.052 & 0.586 & 0.811 & 0.954 & 0.987 \tabularnewline 
        DORN~\cite{fu2018deep} & Depth & 0.115 & 0.051 & 0.509 & 0.828 & 0.965 & 0.992 \tabularnewline 
        \bottomrule
    \end{tabularx}
	\captionof{table}{\textbf{NYU:} Quantitative results on NYU V2~\cite{Silberman:ECCV12} dataset. \textbf{Top -} Our method. \textbf{Bottom -} Supervised methods.}
	\label{table:NYU}
\end{figure*}
For rendered DoF supervision, we consider four datasets~\cite{eigen2014depth,saxena2006learning,Silberman:ECCV12,Carvalho2018eccv3drw} with ground truth depth, where we render focused images with different focus distances. We denote by F1, F2, F6, F10 the four training setups, which differ by the number of rendered focused images used in training. The order in which focal distances are selected, is defined by the following focal sequence $[0.2, 0.8, 0.1, 0.9, 0.3, 0.7, 0.4, 0.6, 0.5, 0.35]$, where each number represents the percent of the maximum depth used for each dataset. For example, F2 employs focal distances of 0.2 and 0.8 times the maximal depth.

We perform two types of evaluations. First, we evaluate our method for each dataset with different numbers of focused images during training, and compare our results with other unsupervised methods, as well as with supervised ones. The evaluation measures are those commonly used in the literature~\cite{Geiger2013IJRR,saxena2006learning, saxena2007learning}
and include various RMSE measures and a thresholded error rate.

Tab.~\ref{table:KITTI} and~\ref{table:Make3D} show that our method outperforms monocular and stereo supervision methods on the KITTI and Make3D dataset. 
This also holds when the previous methods are trained with additional data obtained from the Cityscapes dataset. 
In comparison to the depth supervised methods, we outperform all methods on KITTI, with the exception of~\cite{fu2018deep}, and outperform~\cite{fu2018deep,li2015depth} on Make3D.
In Fig.~\ref{figure:KITTI_Images}, we present qualitative results of our method compared to the state-of-the-art {\em unsupervised} method~\cite{wang2018learning} on the KITTI dataset.
As can be seen in Tab.~\ref{table:NYU}, there are no literature unsupervised methods reported for the NYU dataset, where we are slightly outperformed by the supervised methods. 

We next preform cross domain evaluation compared to the published models of the state-of-the-art supervised method~\cite{fu2018deep}, where training is performed on KITTI or NYU, and tested on different datasets. These tests are meant to evaluate the specificity of the learned network to a particular dataset.
Since the absolute depth differs between datasets, we evaluate the methods by computing the Pearson correlation metric. 
Results are shown in Tab.~\ref{table:cross}. As can be seen, when transferring from both KITTI and NYU, we outperform the directly supervised method. The gap is especially visible for the NYU network.

\begin{table}[t]
	\small
    \centering
	\begin{tabularx}{\linewidth}{llC}
        \toprule
        Transition  & Algorithm & Correlation\tabularnewline 
        \midrule
        \multirow{3}{*}{KITTI $\rightarrow$ NYU}& DORN~\cite{fu2018deep} & 0.423 $\pm$ 0.010\tabularnewline
        & Ours F1 & 0.121 $\pm$ 0.006\tabularnewline  
        & \textbf{Ours F10} & \textbf{0.429} $\pm$ 0.009\tabularnewline 
        \cmidrule(r{14pt}l{8pt}){1-3}
        \multirow{3}{*}{KITTI $\rightarrow$ Make3D}& DORN~\cite{fu2018deep} & 0.616 $\pm$ 0.011 \tabularnewline
        & Ours F1 & 0.484 $\pm$ 0.019 \tabularnewline 
        & \textbf{Ours F10} & \textbf{0.642} $\pm$ 0.014 \tabularnewline 
        \cmidrule(r{14pt}l{8pt}){1-3}
        \multirow{3}{*}{KITTI $\rightarrow$ D3Net} & DORN~\cite{fu2018deep} & 0.145 $\pm$ 0.048\tabularnewline
        & Ours F1 & 0.148 $\pm$ 0.032 \tabularnewline 
        & \textbf{Ours F10} & \textbf{0.275} $\pm$ 0.054\tabularnewline 
        \midrule
        \multirow{2}{*}{NYU $\rightarrow$ KITTI}& DORN~\cite{fu2018deep} & 0.456 $\pm$ 0.006 \tabularnewline
        & Ours F1 & 0.567 $\pm$ 0.006 \tabularnewline
        & \textbf{Ours F10} & \textbf{0.634} $\pm$ 0.005 \tabularnewline
        \cmidrule(r{14pt}l{8pt}){1-3}
        \multirow{2}{*}{NYU $\rightarrow$ Make3D}& DORN~\cite{fu2018deep} & 0.250 $\pm$ 0.019\tabularnewline
        & Ours F1 & 0.249 $\pm$ 0.032 \tabularnewline
        & \textbf{Ours F10} & \textbf{0.456} $\pm$ 0.022 \tabularnewline
        \cmidrule(r{14pt}l{8pt}){1-3}
        \multirow{3}{*}{NYU $\rightarrow$ D3Net} & DORN~\cite{fu2018deep} & 0.260 $\pm$ 0.054\tabularnewline
        & \textbf{Ours F1} & \textbf{0.530} $\pm$ 0.048\tabularnewline 
        & Ours F10 & 0.434 $\pm$ 0.052\tabularnewline
        \bottomrule
	\end{tabularx}
	\caption{Quantitative results for cross domain evaluation. Models are trained on domain A and tested on domain B. 
    Reported numbers are mean $\pm$ standard error.
    }
	\label{table:cross}
\end{table}

We also provide cross-domain results for the outdoor images of the DSLR dataset, where no ground truth depth is provided, using the PSNR and SSIM metrics. Tab.~\ref{table:DSLR} shows in this case that our method transfers better from NYU and only slightly better from KITTI in comparison to~\cite{fu2018deep}. 
\begin{table}[t]
	\small
	\centering
	\begin{tabularx}{\linewidth}{llCC}
        \toprule
        Transition & Algorithm  & PSNR & SSIM \tabularnewline 
        \midrule
        \multirow{3}{*}{KITTI $\rightarrow$ DSLR} & DORN~\cite{fu2018deep} & 24.95 & 0.823 \tabularnewline
        & Ours F1 & 24.91 & 0.822 \tabularnewline 
        & \textbf{Ours F10} & \textbf{24.98} & \textbf{0.826} \tabularnewline 
        \midrule
        \multirow{3}{*}{NYU $\rightarrow$ DSLR} & DORN~\cite{fu2018deep} & 24.73 & 0.749 \tabularnewline
        & Ours F1 &  24.97 & {\bf 0.774} \tabularnewline 
        & \textbf{Ours F10} & \textbf{24.97} & 0.773 \tabularnewline 
        \bottomrule
	\end{tabularx}
	\caption{Quantitative results on the outdoor DSLR~\cite{Carvalho2018eccv3drw} test set, reported as mean value of PSNR and SSIM of the reconstructed focused image.}
	\label{table:DSLR}
    \begin{tabularx}{\linewidth}{lCCCC}
        \toprule
        Model & F1 & F2  & F6 & F10\tabularnewline 
        \midrule
        ASPP & 5.412 & 4.422 & 4.311 & 4.194\tabularnewline
        ASPP + D  & 5.285 & 4.351 & 4.170 & 4.190\tabularnewline
        ASPP + SA & 5.387 & 4.402 & 4.232 & 4.188\tabularnewline
        \textbf{Our} & \textbf{5.187} & \textbf{4.233} & \textbf{4.144} & \textbf{4.186}\tabularnewline 
        \bottomrule
	\end{tabularx}
	\caption{A comparison on KITTI between the original ASPP and our dense ASPP with self-attention. We denote `D' for Dense connections and `SA' for Self-Attention. RMSE is shown for focused image stacks of different sizes.}
	\label{table:compare}
	
    \setlength{\tabcolsep}{6pt} 
    \renewcommand{\arraystretch}{1} 
    \setlength{\thickmuskip}{0mu}
	\small
    \begin{tabular}{@{}l@{~~~}c@{~~}c@{~~}c@{~~~~~}c@{~~}c@{~~}c@{}}
        \toprule
        \multirow{2}{*}{Rendering} & \multicolumn{3}{c}{F1} & \multicolumn{3}{c}{F2}\\
         & Abs Rel & RMSE & $\delta<1.25$ & Abs Rel & RMSE & $\delta<1.25$\\
        \midrule
        \cite{srinivasan2018aperture} & 0.489 & 12.395 & 0.293 & 0.636 & 11.177 & 0.230 \\
        \cite{srinivasan2018aperture}+BF & 0.379 & 11.921 & 0.354 & 0.339 & 11.612 & 0.418 \\
        \textbf{Ours} & \textbf{0.141} & \textbf{5.187} & \textbf{0.846} & \textbf{0.129} & \textbf{4.233} & \textbf{0.856} \\
        \bottomrule
	\end{tabular}
    \captionof{table}{{\color{black}A comparison on KITTI dataset between different blur methods on top of our network. BF= bilateral filtering.}}
    \label{tbl:blurlayer}
    
    \small
    \begingroup
    \setlength{\tabcolsep}{2pt} 
    \renewcommand{\arraystretch}{1} 
    \begin{tabular}{cc}
    	\includegraphics[width=0.48\linewidth, height=0.4\linewidth]{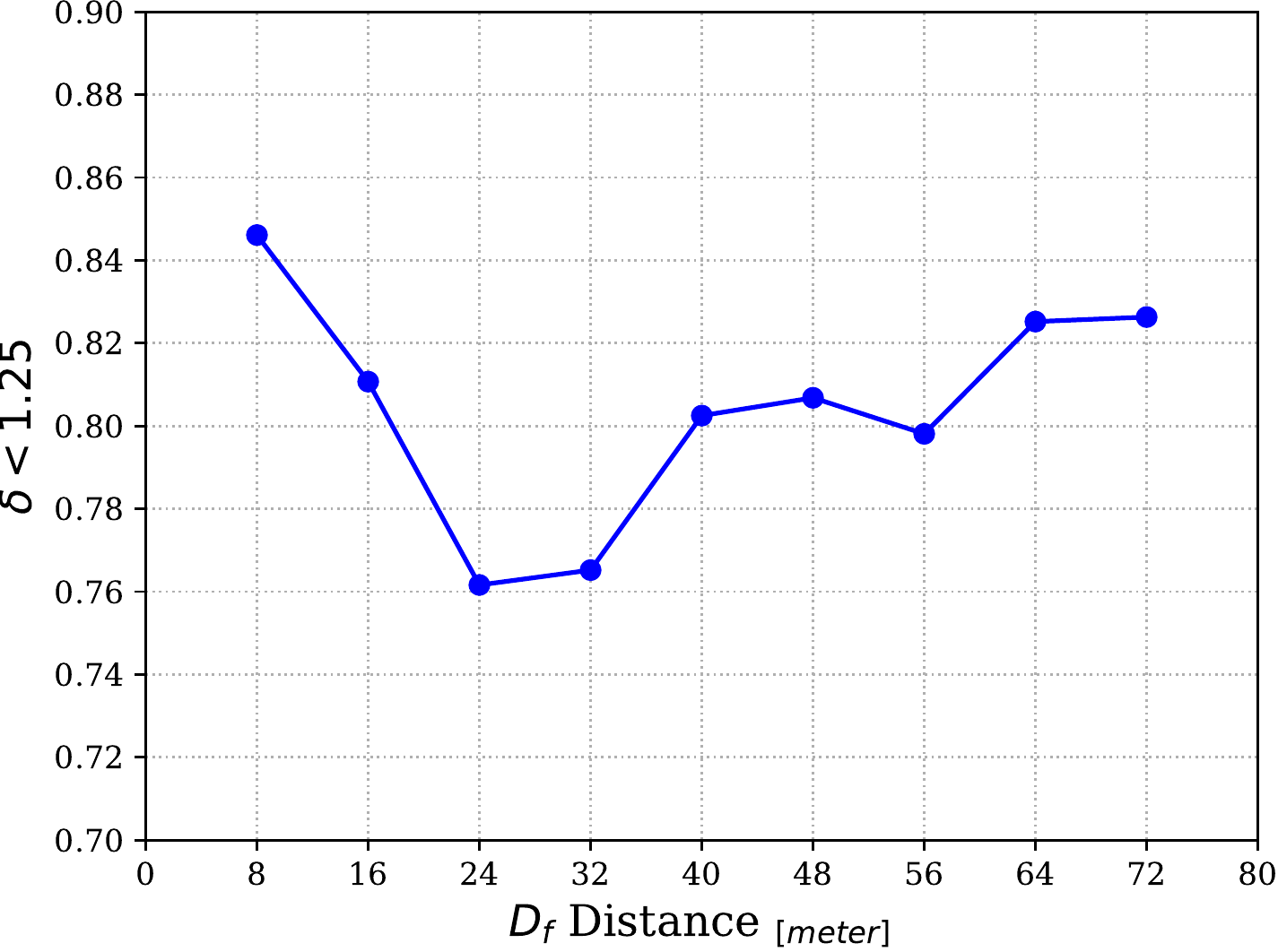} &
		\includegraphics[width=0.48\linewidth, height=0.4\linewidth]{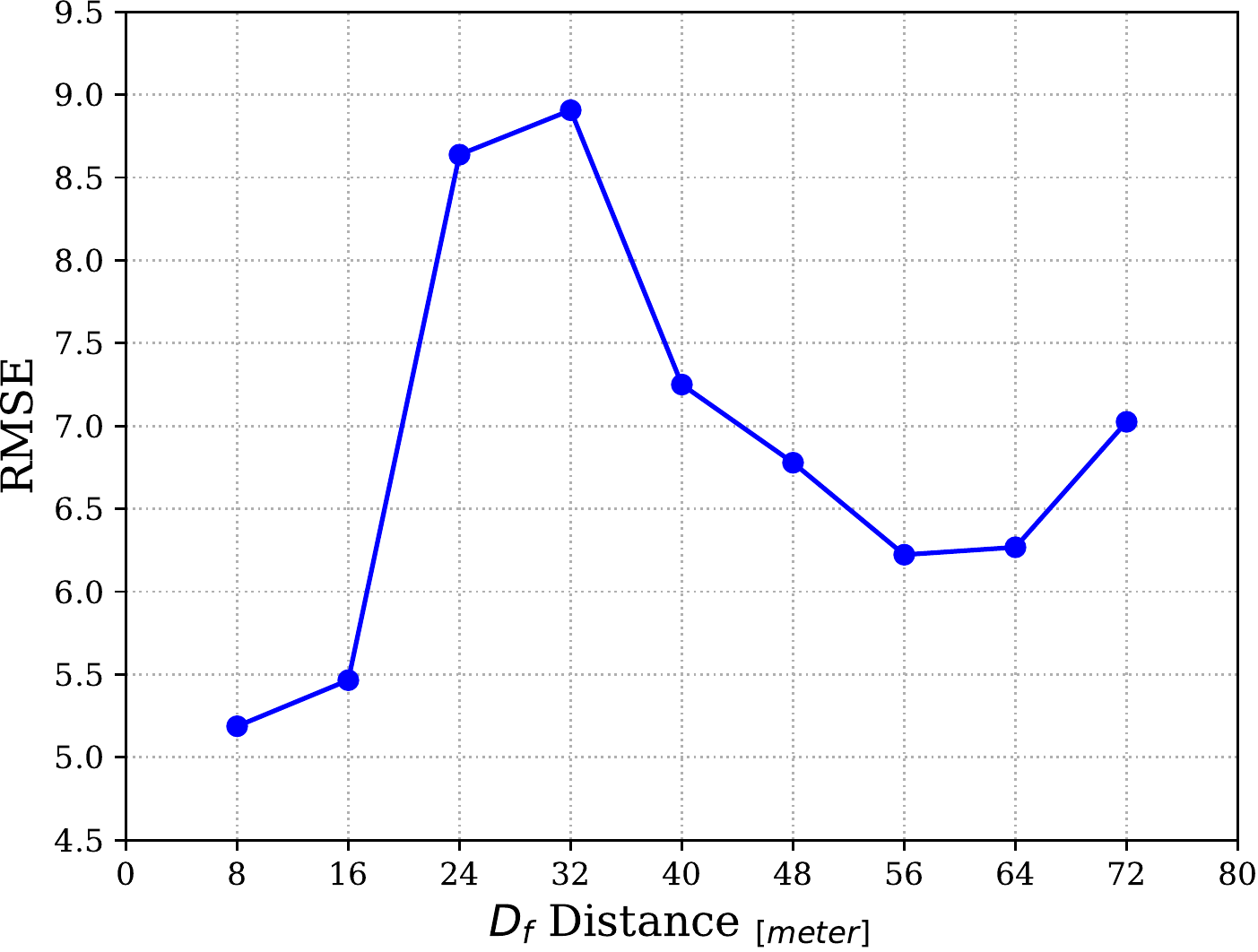}\\
    	(a) & (b)
    \end{tabular}
    \endgroup
    \captionof{figure}{(a) $\delta<1.25$, higher is better, for training F1 with different focus distance. (b) RMSE, lower is better.}
    \label{fig:graphs}
    
\end{table}

\subsection{Ablation Studies}
\noindent\textbf{The Effect of Focal Distance}\quad
\label{sec:depth_ambiguity}
Because the focus distance $D_f$ and DoF range are positively correlated, training with a far focus distance increases the DoF and puts a large range of distances in focus. As a result, focus cues are lowered, causing performance to decrease.
In Fig.~\ref{fig:graphs} we present, for the Make3D dataset, the accuracy of F1 training with different focus distances, where a clear decrease in performance is seen at mid-range $D_f$ and an increase afterward, as a result of the dataset maximum depth, capping the far DoF distance, \ie lowering the DoF range, and increasing focus cues for closer objects.

\noindent\textbf{Dense ASPP with Self-Attention}\quad
We evaluate our dense ASPP with self-attention in comparison to three versions of the original ASPP model: vanilla ASPP, ASPP with dense connections and ASPP with self-attention. In order to differentiate between different ambiguity scenarios, {training is preformed with the F1, F2, F6 and F10 methods. As can be seen in Tab~\ref{table:compare}, our model outperform the different ASPP versions. However, as the number of focused images increases, the gaps are reduced.}

\noindent\textbf{Different rendering methods}\quad
To further compare with~\cite{srinivasan2018aperture}, we have conducted a test on the KITTI dataset, where we replaced our rendering network $g$ with their compositional rendering, and modified our depth network $f$'s last layer to output 80 depth probabilities (similar to~\cite{srinivasan2018aperture}). From Tab.~\ref{tbl:blurlayer}, the compositional method of~\cite{srinivasan2018aperture} preforms poorly on KITTI in the F1 and F2 setting.

\section{Conclusion}
We propose a method for learning to estimate depth from a single image, based on focus cues. Our method outperforms the similarly supervised method~\cite{srinivasan2018aperture} and all other unsupervised literature methods. In most cases, it matches the performance of directly supervised methods, when evaluated on test images from the training domain. Since focus cues are more generic than content cues, our method outperforms the state-of-the-art supervised method in cross domain evaluation on all available literature  datasets.

We introduce a differentiable PSF convolutional layer, which propagates image based losses back to the estimated depth. We also contribute a new architecture that introduces dense connection and Self-Attention to the ASPP module. 
Our code is available as part of the supplementary material, and on GitHub \url{https://github.com/shirgur/UnsupervisedDepthFromFocus}.

\section*{Acknowledgment}
This project has received funding from the European Research Council (ERC) under the European Unions Horizon 2020 research and innovation programme (grant ERC CoG 725974). The contribution of the first author is part of a Ph.D. thesis research conducted at Tel Aviv University.

{\small
\bibliographystyle{ieee}
\bibliography{allbib}
}

\clearpage
\title{Supplementary: Differentiable Gaussian PSF Layer}
\author{}
\maketitle

\begin{figure}
    \centering
    \includegraphics[width=\linewidth]{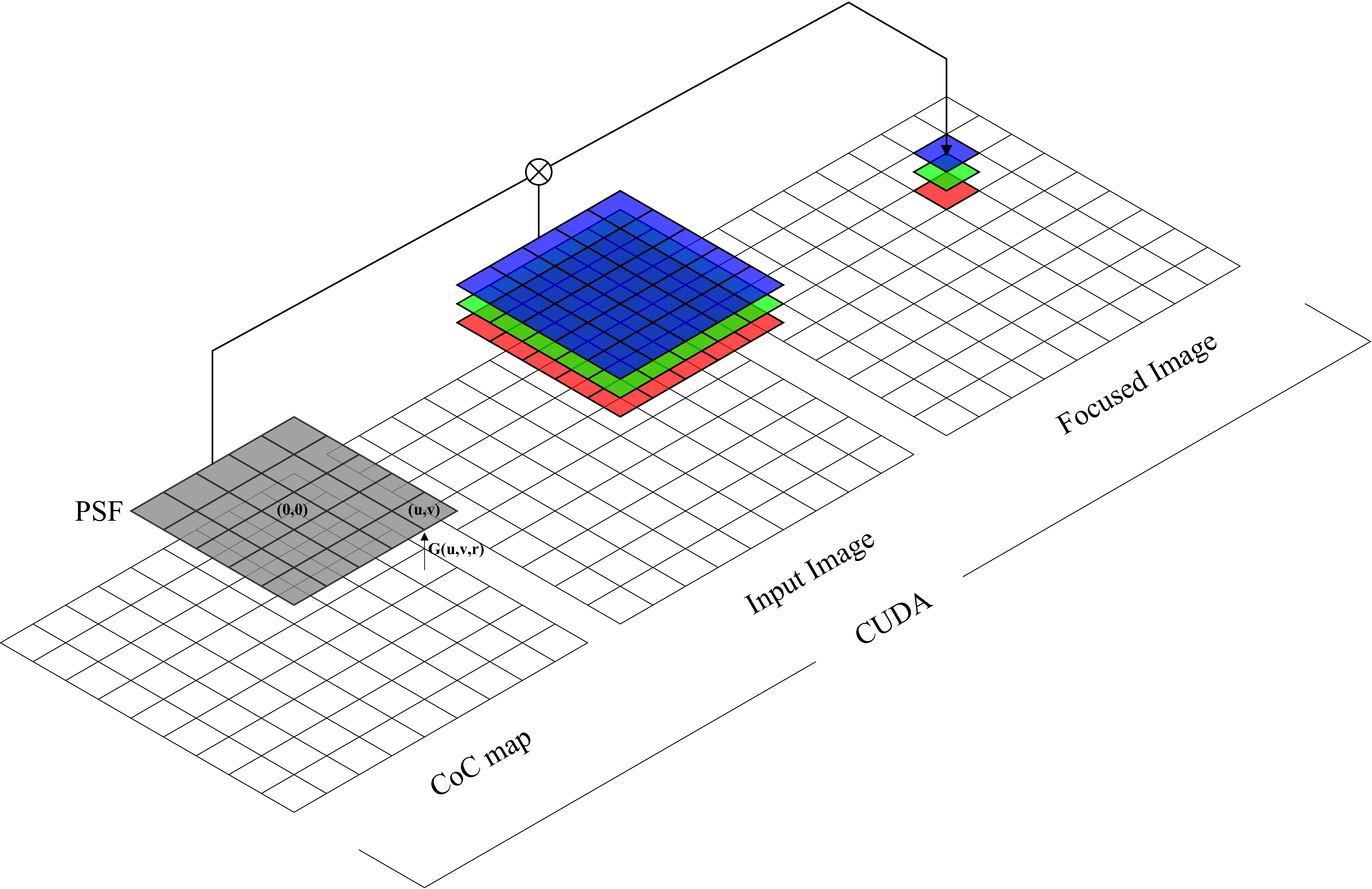}
    \caption{Illustration of the PSF convolution layer for focus rendering. each kernel is spatially computed from the CoC map and multiplied with the corespondent image patch both channel and element wise.}
    \label{fig:PSF}
\end{figure}
\setcounter{equation}{0}
In this appendix we show the mathematical development of our PSF layer for backward and forward pass.
For easy notation we define:
\[
	M_{x-i, y-j} := M_[x-i, y-j]
\]
where $M$ is any matrix or function, $(x, y)$ are location indices and $(u,v)$ are offsets.
An illustration of the the PSF convolution is shown in Fig.~\ref{fig:PSF}.

\noindent\textbf{Forward Pass:} For a PSF kernel of size $m \times m$ where $m$ is odd, we consider an offsets set $\Omega$ of the form:
\[
	\Omega := \left\lbrace (u,v):\;u,v\in[-\frac{m}{2},\dots,0,\dots,\frac{m}{2}]\in\mathbb{N} \right\rbrace 
\]
The forward pass is calculate by the following equations, where $I$ is the input image, $J$ is the output image and $C$ is the CoC map:
\begin{align}
	\mathcal{F}_{x,y}(u, v) &= \frac{2}{\pi C_{x,y}^2}\exp\bigg(-2\bigg(\frac{u^2+ v^2}{C_{x,y}^2}\bigg)\bigg) \\[10pt]
	J_{x, y} :&= (I \circledast F) \\[10pt]
    \nonumber
    &= \frac{\int\limits_{u,v \in \Omega} I_{x-u, y-v}\mathcal{F}_{x-u, y-v}(u, v)du dv}{\int\limits_{u',v' \in \Omega} \mathcal{F}_{x-u', y-v'}(u', v')du' dv'}
\end{align}

\noindent\textbf{Backward Pass:} 
We first compute derivative of loss $\mathcal{L}$ w.r.t. image $I$ at location $(x,y)$:
\begin{align}
\bigg(\frac{\partial \mathcal{L}}{\partial I_{x,y}}\bigg) &= \int\limits_{\substack{u,v \in \Omega\\s=x-u\\t=y-v}} \bigg(\frac{\partial \mathcal{L}}{\partial J_{s,t}}\bigg)\bigg(\frac{\partial J_{s,t}}{\partial I_{x,y}}\bigg)dudv\\[10pt]
\bigg(\frac{\partial J_{s,t}}{\partial I_{x,y}}\bigg) &= \frac{\mathcal{F}_{x,y}(u,v)}{\int\limits_{u',v' \in \Omega} \mathcal{F}_{s-u',t-v'}(u', v')du' dv'}
\end{align}

Secondly compute derivative of loss $\mathcal{L}$ w.r.t. CoC map $C$ at location $(x,y)$:
\begin{align}
\bigg(\frac{\partial \mathcal{L}}{\partial C_{xy}}\bigg) &= \int\limits_{\substack{u,v \in \Omega\\s=x-u\\t=y-v}} \bigg(\frac{\partial \mathcal{L}}{\partial J_{st}}\bigg)\bigg(\frac{\partial J_{st}}{\partial C_{xy}}\bigg)dudv\\[10pt]
J_{st} :&= \phi\cdot\psi \\[10pt]
\phi :&= \int_{u',v' \in \Omega} I_{s-u',t-v'}\mathcal{F}_{s-u',t-v'}(u', v')du' dv' \\[10pt]
\psi :&= \frac{1}{\int_{u',v' \in \Omega} \mathcal{F}_{s-u',t-v'}(u', v')du' dv'}
\end{align}
Note that we consider only $C$ where $(s-u', t-v')=(x,y)$, where in these locations the offset for $\mathcal{F}$ is the opposite of $(u,v)$ meaning $(-u, -v)$, but because we only calculate their second power we can consider $\mathcal{F}_{x,y}(u,v)$:
\begin{align}
\bigg(\frac{\partial \phi}{\partial C_{x,y}}\bigg) &= \frac{-2\cdot2I_{x,y}}{\pi C^3_{x,y}}\exp\bigg(-2\bigg(\frac{u^2+ v^2}{C_{x,y}^2}\bigg)\bigg) \\[10pt]
\nonumber
&+ \frac{2\cdot2(u^2+ v^2)}{C^3_{xy}}\frac{2I_{x,y}}{\pi C^2_{xy}}\exp\bigg(-2\bigg(\frac{u^2+ v^2}{C_{x,y}^2}\bigg)\bigg) \\[10pt]
\nonumber
&= I_{x,y}\exp\bigg(-2\bigg(\frac{u^2+ v^2}{C_{x,y}^2}\bigg)\bigg)\frac{2}{\pi C_{x,y}^2}\\[10pt]
\nonumber
&\cdot \frac{4(u^2+ v^2) - 2C_{x,y}^2}{C_{x,y}^3}\\[10pt]
\nonumber
&= I_{x,y} \mathcal{F}_{x,y}(u, v) \frac{4(u^2+ v^2) - 2C_{x,y}^2}{C_{x,y}^3}\\[10pt]
\bigg(\frac{\partial \psi}{\partial C_{x,y}}\bigg) &= -\bigg(\frac{\partial \phi}{\partial C_{x,y}}\bigg)\frac{\psi^2}{I_{x,y}}
\end{align}
Now that we have each individual derivative we can compute according to the \textit{product rule}:
\begin{align}
\bigg(\frac{\partial J_{s,t}}{\partial C_{x,y}}\bigg) &= \phi \bigg(\frac{\partial \psi}{\partial C_{x,y}}\bigg) + \psi \bigg(\frac{\partial \phi}{\partial C_{x,y}}\bigg)\\[10pt]
\phi \bigg(\frac{\partial \psi}{\partial C_{x,y}}\bigg) &= \frac{J_{s,t}}{\psi}\bigg(-\bigg(\frac{\partial \phi}{\partial C_{x,y}}\bigg)\frac{\psi^2}{I_{x,y}}\bigg) \\[10pt]
\nonumber
&= -\frac{J_{s,t}\psi}{I_{x,y}}\bigg(\frac{\partial \phi}{\partial C_{x,y}}\bigg) \\[10pt]
\bigg(\frac{\partial J_{s,t}}{\partial C_{x,y}}\bigg) &= -\frac{J_{s,t}\psi}{I_{x,y}}\bigg(\frac{\partial \phi}{\partial C_{x,y}}\bigg) + \psi \bigg(\frac{\partial \phi}{\partial C_{x,y}}\bigg) \\[10pt]
\nonumber
&= \frac{\psi}{I_{x,y}} \bigg(\frac{\partial \phi}{\partial C_{x,y}}\bigg)(I_{x,y}-J_{s,t})
\end{align}
Finally we get the complete answer:
\begin{align}
\xi_{x,y}(u, v) &:= \frac{4(u^2+ v^2) - 2C_{x,y}^2}{C_{x,y}^3}\\[10pt]
\bigg(\frac{\partial J_{s,t}}{\partial C_{x,y}}\bigg) &= \frac{\psi}{I_{x,y}}I_{x,y} \mathcal{F}_{x,y}(u, v)\xi_{x,y}(u, v) (I_{x,y}-J_{s,t})\\[10pt]
\nonumber
&= \frac{\xi_{x,y}(u, v) (I_{x,y}-J_{s,t})\mathcal{F}_{x,y}(u, v)}{\int_{u',v' \in \Omega} \mathcal{F}_{s-u',t-v'}(u', v')du' dv'}
\end{align}
\end{document}